\documentclass[conference]{IEEEtran}
\IEEEoverridecommandlockouts
\usepackage{cite}
\usepackage{amsmath,amssymb,amsfonts}
\usepackage{algorithmic}
\usepackage{graphicx}
\usepackage{textcomp}
\usepackage{xcolor}
\usepackage{listings}
\usepackage{enumitem}
\usepackage{tabularx}
\usepackage{booktabs}
\usepackage{multirow}
\newcolumntype{C}{>{\centering\arraybackslash}X}
\def\BibTeX{{\rm B\kern-.05em{\sc i\kern-.025em b}\kern-.08em
    T\kern-.1667em\lower.7ex\hbox{E}\kern-.125emX}}
\begin{document}

\definecolor{lightgray}{rgb}{0.95, 0.95, 0.95}
\definecolor{darkgray}{rgb}{0.4, 0.4, 0.4}
\definecolor{purple}{rgb}{0.65, 0.12, 0.82}

\lstdefinestyle{jsonstyle}{
    backgroundcolor=\color{lightgray},
    basicstyle=\ttfamily\small,
    breakatwhitespace=false,         
    breaklines=true,                 
    captionpos=b,                    
    keepspaces=true,                 
    numbers=none,            
    showspaces=false,                
    showstringspaces=false,
    showtabs=false,                  
    tabsize=2,
    stringstyle=\color{purple},
    commentstyle=\color{darkgray},
    frame=single             
}

\definecolor{codebg}{rgb}{0.95,0.95,0.95}
\definecolor{keywordcolor}{rgb}{0.13,0.13,0.8}
\definecolor{commentcolor}{rgb}{0.0,0.5,0.0}

\lstdefinelanguage{PDDL}{
  morekeywords={
    define, domain, problem, :requirements, :types, :constants, 
    :predicates, :action, :parameters, :precondition, :effect, 
    :init, :goal, and, or, not, imply, exists, forall, when
  },
  sensitive=false,       
  morecomment=[l]{;},    
  morestring=[b]"
}

\lstdefinestyle{pddlstyle}{
    language=PDDL,                  
    backgroundcolor=\color{codebg},
    basicstyle=\ttfamily\small,
    keywordstyle=\color{keywordcolor}\bfseries,
    commentstyle=\color{commentcolor}\itshape,
    breaklines=true,
    captionpos=b,
    numbers=none,
    numbersep=5pt,
    frame=single
}

\title{DUPLEX: Agentic Dual-System Planning via LLM-Driven Information Extraction\\
\thanks{$^{\dagger}$Contributed equally, $^{*}$Corresponding authors}
}

\author{
  \IEEEauthorblockN{Keru Hua$^{\dagger}$}
  \IEEEauthorblockA{\textit{Midea Corporate Research Center} \\
\textit{Midea Group}\\
Foshan, China \\
huakr@midea.com}
  \\[0.1cm] 
  \IEEEauthorblockN{Yaoying Gu}
  \IEEEauthorblockA{\textit{Midea Corporate Research Center} \\
\textit{Midea Group}\\
Foshan, China \\
guyy30@midea.com}
  
  \and
  
  \IEEEauthorblockN{Ding Wang$^{\dagger,*}$}
  \IEEEauthorblockA{\textit{Midea Corporate Research Center} \\
\textit{Midea Group}\\
Foshan, China \\
ding2.wang@midea.com}
  \\[0.1cm]
  \IEEEauthorblockN{Xiaoguang Ma$^{*}$}
  \IEEEauthorblockA{\textit{Robot Science and Engineering Department} \\
\textit{Northeastern University}\\
Shenyang, China \\
maxg@mail.neu.edu.cn}
}



\maketitle
\begin{abstract}
While Large Language Models (LLMs) provide semantic flexibility for robotic task planning, their susceptibility to hallucination and logical inconsistency limits their reliability in long-horizon domains. To bridge the gap between unstructured environments and rigorous plan synthesis, we propose DUPLEX, an agentic dual-system neuro-symbolic architecture that strictly confines the LLM to schema-guided information extraction rather than end-to-end planning or code generation. In our framework, a feed-forward Fast System utilizes a lightweight LLM to extract entities, relations etc. from natural language, deterministically mapping them into a Planning Domain Definition Language (PDDL) problem file for a classical symbolic planner. To resolve complex or underspecified scenarios, a Slow System is activated exclusively upon planning failure, leveraging solver diagnostics to drive a high-capacity LLM in iterative reflection and repair. Extensive evaluations across 12 classical and household planning domains demonstrate that DUPLEX significantly outperforms existing end-to-end and hybrid LLM baselines in both success rate and reliability. These results confirm that The key is not to make the LLM plan better, but to restrict the LLM to the part it is good at—structured semantic grounding—and leave logical plan synthesis to a symbolic planner.
\end{abstract}


\section{Introduction}

Deploying embodied agents in open-world environments requires robust, long-horizon planning under strict physical constraints. Classical symbolic planners, built upon formal languages like the Planning Domain Definition Language (PDDL), have traditionally served as rigorous solvers for such tasks. These engines guarantee deterministic executability, mathematically verifiable safety, and optimal search \cite{mcdermott1998pddl, fox2003pddl21, ghallab2004automatedplanning, hoffmann2001ff, helmert2006fast, howey2004val}. However, they suffer from a severe \emph{knowledge acquisition bottleneck}: translating high-dimensional, unstructured environment states and human intents into rigid symbolic representations requires extensive manual engineering. This inflexibility severely limits their applicability in dynamic, real-world scenarios.

To overcome this bottleneck, recent approaches have turned to Large Language Models (LLMs) to leverage their exceptional commonsense reasoning and instruction-following capabilities \cite{brown2020language, ouyang2022instructgpt, openai2023gpt4, huang2022language, ahn2022i, driess2023palme, liang2023code}. Yet, applying LLMs to planning introduces critical vulnerabilities. End-to-end long-horizon planning based on LLM remains fundamentally unreliable, such as frequently hallucinating actions, losing long-range consistency, and violating implicit physical preconditions \cite{valmeekam2022large, pallagani2023understanding}. It is often very time-consuming too. All of this makes LLM-as-Planner unsuitable for safety-critical tasks that demand certainties. 

On the other hand, neuro-symbolic translation methods (such as LLM+P\cite{liu2023llm+}) attempt to use the LLM to directly write formal PDDL code. While this bridges the gap between the real-world environment and PDDL representations, and mitigates the knowledge acquisition bottleneck, it forces the LLM to generate perfectly accurate PDDL code — demanding not only flawless syntax, but also the exact, exhaustive formulation of all relevant objects, predicates, and state conditions etc. without any omissions or hallucinations. Such strict generation is highly brittle and prone to compilation and runtime failures in complex, long-horizon planning problems.

We therefore employ LLM as an \emph{Information Extractor (IEr)} guided by schemas, rather than a end-to-end planner or code generator. By reducing the complex planning or code generating challenge to Information Extraction (IE) problem, which is a classical Natural Language Processing (NLP) problem, the task becomes significantly simpler and more suitable for LLMs. Consequently, even smaller-scale LLMs may perform well. After information extraction, a deterministic pre-programmed mapper function then robustly translates the information into PDDL file, which is subsequently solved by a symbolic planner \cite{helmert2006fast, howey2004val}. We call this efficient, feed-forward pipeline the \textbf{Fast System}.

To guarantee high success rates in complex scenarios, we further propose \textbf{DUPLEX}, an agentic dual-system architecture in which a complementary \textbf{Slow System} is built upon the Fast System. Activated only when the Fast System fails, the Slow System serves as a robust fallback agent. Upon failure, the error messages from the symbolic planner are routed to a larger, more capable LLM, which analyzes and formulates repair strategies, and iteratively modifies the PDDL file until the planner successfully yields a solution. By confining intensive reasoning and self-correction only to failure cases, DUPLEX enables recovery from semantic and logical errors without incurring heavy computational overhead on routine tasks, complementing recent progress in reflective language agents \cite{shinn2023reflexion, madaan2023selfrefine, dhuliawala2023cove, yao2023tot}.

Our core contributions are:
\begin{itemize}
    \item \textbf{A Novel LLM-Symbolic Pipeline:} the planning problem is solved by a process of "information extraction $\rightarrow$ converting to PDDL$\rightarrow$ plan generation", and a lightweight LLM is employed as the information extractor.
    \item \textbf{The DUPLEX Architecture:} an agentic dual-system architecture is proposed, which incorporates reflection mechanism, powered by a larger LLM.
    \item \textbf{Empirical Validation:} experiments validate our approach on 12 domains, demonstrating substantial improvements in reliability and efficiency.
\end{itemize}

\section{Related Work}
This section provides a brief overview of prior task planning works: (i) classical symbolic planning, (ii) LLM-as-planner, (iii) LLM+P

\subsection{Classical Symbolic Planning}
Symbolic planning based on PDDL remains a foundational methodology for governing autonomous behaviors in embodied agents \cite{mcdermott1998pddl, fox2003pddl21}. Provided with rigorously specified domain and problem PDDL files, modern heuristic planners like Fast Downward leverage deterministic search algorithms to generate mathematically verifiable plans \cite{helmert2006fast, hoffmann2001ff}. Particularly in complex, long-horizon scenarios, these explicit representations offer strict controllability. Furthermore, they allow independent validators to check that all preconditions, transition effects, and goal specifications etc. are satisfied\cite{howey2004val, ghallab2004automatedplanning}.

\noindent\textbf{Limitations:} The primary bottleneck lies not in the solver itself, but in the semantic grounding between the physical world and PDDL representations. Domain experts need to manually map unstructured observations into PDDL representation, this "translation" work is notoriously brittle, laborious and expensive to scale.

\subsection{LLM-as-Planner}
Recent literature extensively explores the \emph{LLM-as-Planner} paradigm, wherein foundation models directly convert natural language instructions into executable action sequences or task decompositions \cite{huang2022language, song2023llm}. These approaches excel at zero-shot generalization by leveraging LLMs' rich commonsense, which is very difficult to manually encode.

Moving beyond text-to-action mapping, contemporary embodied systems integrate LLMs with learned robotic affordances, code generation, and multimodal grounding \cite{ahn2022i, liang2023code, driess2023palme, shah2023lm, wang2023voyager}. These end-to-end generative methods demonstrate impressive adaptability.

\noindent\textbf{Limitations:} When tasked with end-to-end plan generation, LLMs often produce superficially plausible sequences that violate implicit physical preconditions, omit critical intermediate steps, or hallucinate ungrounded actions etc.\cite{valmeekam2022large}. 

\subsection{LLM+P}
A seminal approach bridging language models and symbolic planner is LLM+P \cite{liu2023llm+}, which explicitly decouples problem formulation from algorithmic search. In this framework, an LLM is provided with a predefined domain PDDL file and is prompted to convert a problem description in natural language to a PDDL file. This problem PDDL file is then processed by a symbolic planner to compute an optimal plan.

\noindent\textbf{Limitations:} While LLM+P leverages symbolic solvers to guarantee feasible/optimal plans, in complex scenarios, LLM may frequently generate syntactically invalid or semantically incorrect PDDL file. Since the framework lacks a closed-loop feedback mechanism, any mistake in "translation" highly possibly causes the entire pipeline fail.

\section{Preliminaries}
This section outlines the core concepts utilized in this work, including the formal notation for classical planning problems, LLM-driven Information Extraction, and iterative self-correction mechanisms.

\subsection{The Classical Planning Problem}
Formally, a classical planning problem $\mathcal{P}$ is defined by a tuple:
$$ \mathcal{P} = \langle \mathcal{S}, \mathcal{A}, \mathcal{T}, s_{\text{init}}, \mathcal{S}_{\text{G}} \rangle $$
where:
\begin{itemize}
    \item $\mathcal{S}$ is a finite, discrete set of states, i.e. state space. Each state $s \in \mathcal{S}$ is defined by the values of a fixed set of state variables.
    
    \item $\mathcal{A}$ is a finite set of symbolic actions.
    
    \item $\mathcal{T}: \mathcal{S} \times \mathcal{A} \rightarrow \mathcal{S}$ is the state transition function. Given a current state and an action, it deterministically outputs the next state.
    
    \item $s_{\text{init}} \in \mathcal{S}$ is the initial state.
    
    \item $\mathcal{S}_{\text{G}} \subseteq \mathcal{S}$ is the set of goal states. 
\end{itemize}

A solution to a planning problem is a sequence of actions $\pi = \langle a_1, \dots, a_n \rangle, a_i\in \mathcal{A}$. $\pi$ is executable from the initial state $s_{\text{init}}$, such that the resulting final state $s_n = \mathcal{T}(s_{\text{init}}, \pi) \subseteq \mathcal{S}_{\text{G}}$. The executability requirement means that for all $i \in \{1, \dots, n\}$, the preconditions of action $a_i$ must hold in state $s_{i-1}$.

A planning problem $\mathcal{P}$ is often formalized using languages like PDDL, where a \textbf{Domain PDDL file} defines types and the actions ($\mathcal{A}$) with their preconditions and effects, and a \textbf{Problem PDDL file} specifies the objects, the initial state ($s_{\text{init}}$), and the goal conditions ($\mathcal{S}_{\text{G}}$).

\subsection{LLM-Driven Information Extraction}
Information Extraction (IE) is a foundational Natural Language Processing (NLP) task focused on extracting structured knowledge—such as entities, relations, and attributes—from unstructured text. While recent studies highlight that Large Language Models (LLMs) frequently struggle with the strict, multi-step logical deduction required for long-horizon planning \cite{valmeekam2022large}, they have demonstrated exceptional and robust performance in classic IE tasks, including Named Entity Recognition (NER) and Relation Extraction (RE) \cite{xu2024largelanguagemodelsgenerative, wei2024chatiezeroshotinformationextraction}. Rather than forcing the LLM to act as an end-to-end planner or a code generator, our framework relies on this empirical strength. By reframing the interpretation of environment observations and human instructions strictly as an IE problem, we effectively bridge unstructured real-world inputs with the deterministic semantic space required for symbolic planning.

\subsection{LLM Reflexion and Self-Correction}

Recent advancements in autonomous agents have introduced the concept of \textit{reflexion} or self-correction, enabling LLMs to iteratively refine their outputs based on external feedback \cite{shinn2023reflexionlanguageagentsverbal, madaan2023selfrefineiterativerefinementselffeedback}.  Unlike standard one-pass generation, a reflective LLM agent utilizes diagnostic signals—such as compilation failures, execution trace errors, or environment states—to identify root causes and propose targeted repairs. In neuro-symbolic planning, end-to-end LLM generation often suffers from logical inconsistencies. However, by leveraging the deterministic error messages generated by a symbolic planner as highly reliable external grounding signals, an LLM can effectively correct semantic or logical omissions in its generated formulations without requiring human intervention. This iterative refinement paradigm forms the theoretical foundation for the reactive repair mechanism in our proposed architecture.

\section{The Fast System$\colon$ Planning Via Information Extraction}

This section introduces the Fast System. We outline the motivation and detail its three-stage pipeline.

\subsection{Motivation}

Despite showing impressive generalization, in practice the \emph{LLM-as-Planner} paradigm is computationally expensive and rarely succeeds in a single pass on long-horizon tasks. As inherently statistical models, LLMs are less suited for rigorous multi-step reasoning and planning, frequently hallucinating actions and violating complex logical and physical constraints.

Alternatively, by offloading the search process to a symbolic solver, the \emph{LLM+P} paradigm simplifies the LLM's role to converting natural language to PDDL—a deceptively simple task, yet one that becomes fragile in complex domains. Minor semantic omissions or logical inaccuracies in the generated PDDL file frequently cause the solver to fail. Consequently, prompt engineering devolves into a frustrating cycle, where correcting one error often introduces new ones.

To maximize planning success rates, we propose strictly confining the LLM to tasks that remain intractable for traditional methods. We decompose the end-to-end planning process into a streamlined pipeline: \textit{unstructured text input $\rightarrow$ Information Extraction (IE) $\rightarrow$ deterministic PDDL mapping $\rightarrow$ symbolic solving}. Since the PDDL conversion step essentially functions as template filling (illustrated in Listings~\ref{lst:json_extract} and \ref{lst:pddl_gen})
—a task that can be finished by predefined scripts—parsing unstructured text remains the only operation that traditional methods struggle with. Therefore, we exclusively assign the IE task to the LLM. 

By relieving the language model of both complex logical search and strict syntactic generation, this architecture not only improves overall system robustness and success rates, but also enables the deployment of smaller, more efficient LLMs, thereby significantly reducing computational overhead and latency.

\begin{figure}[htbp]
\small 
\centering
\noindent
\begin{minipage}[t]{0.48\textwidth}
\begin{lstlisting}[
  style=jsonstyle, 
  caption={Extracted Information in JSON},
  label={lst:json_extract} % <--- 添加 label
]
{
  "objects": [
    {"id": "apple_01", "type": "apple"},
    {"id": "table_main", "type": "table"},
    {"id": "plate_01", "type": "plate"}
  ],
  "relations": {
    "init": [
      {"subject": "apple_01", 
       "predicate": "on", 
       "object": "table_main"}
    ],
    "goal": [
      {"subject": "apple_01", 
       "predicate": "on", 
       "object": "plate_01"}
    ]
  }
}
\end{lstlisting}
\end{minipage}
\hfill
\begin{minipage}[t]{0.48\textwidth}
\begin{lstlisting}[
  style=pddlstyle, 
  caption={Generated PDDL},
  label={lst:pddl_gen} % <--- 添加 label
]
(:objects
  apple_01 - apple
  table_main - table
  plate_01 - plate
)
(:init
  (on apple_01 table_main)
)
(:goal
  (and 
    (on apple_01 plate_01)
  )
)
\end{lstlisting}
\end{minipage}
\end{figure}

\subsection{The Fast System}

\begin{figure*}[t]
    \centering
    \includegraphics[width=0.9\textwidth]{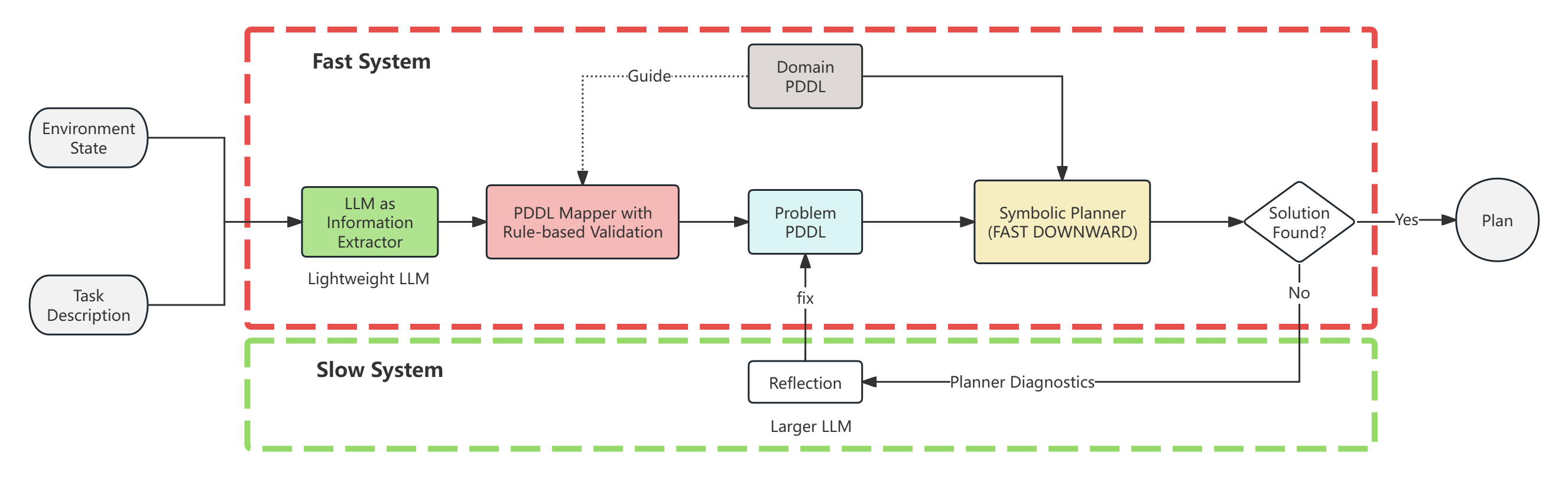}
    \caption{Overview of \textbf{DUPLEX} architecture. \textbf{Fast System}: Unstructured inputs are parsed by a lightweight LLM acting solely as an information extractor. The resulting structured data are deterministically mapped to a problem PDDL file, which is then paired with a domain PDDL file and solved by a symbolic planner. \textbf{Slow System}: Upon planning failure, the resulting planner diagnostics trigger the Slow System, where a larger LLM reflects on the error messages to repair the problem PDDL file, enabling iterative replanning.\vspace{-1em}}
    \label{fig:system}
\end{figure*}

The red bounding box in Figure~\ref{fig:system} illustrates the architecture of the Fast System. Given unstructured task description and environment observation, the Fast System executes a three-stage pipeline: (1) \textbf{Information Extraction}, where an LLM parses unstructured input to identify both explicit and implicit objects, relations, and states; (2) \textbf{Mapping Information to PDDL}, where a pre-programmed script converts the structured data into a PDDL problem file; and (3) \textbf{Symbolic Planner}, where a classical planner leverages a predefined domain PDDL file to compute an optimal action sequence.

\subsection*{Step 1: Guided Information Extraction}

To mitigate hallucination, we prompt the LLM to act strictly as an information extractor, guided by a predefined schema derived from the given domain PDDL file, which outlines all valid object types and predicates etc. The model outputs a structured format (e.g. JSON) capturing three key components:
\begin{itemize}
    \item \textbf{Objects:} The set of distinct entities identified in the environment. These serve as the foundational types for PDDL object declaration.
    \item \textbf{Relations:}  Directed interactions between objects that function as grounded predicates. Each relation specifies a predicate name and its associated object arguments, ready for instantiation in the planning domain.
    \item \textbf{States:} Temporal snapshots of relational facts, including \textit{initial state}, representing the current environmental configuration, and the \textit{goal state}, specifying the target configuration required to satisfy the task objective. 
\end{itemize}

\subsection*{Step 2: Mapping to PDDL with syntactic validation}

The second stage bridges the semantic output of the LLM with the rigid syntactic requirements of symbolic planning. Rather than relying on the LLM for code generation, this step employs a deterministic mapping function. A dedicated script ingests the structured data from the previous step, performs a rule-based validation, and populates standard PDDL templates (illustrated in Listings~\ref{lst:json_extract} and \ref{lst:pddl_gen}). By completely isolating the LLM from PDDL syntax generation, our approach guarantees $100\%$ syntactic correctness, even with lightweight LLMs. This design eliminates the compilation failures—such as missing parentheses or type mismatches—that frequently plague direct LLM-to-PDDL generation methods.

\subsection*{Step 3: Symbolic Plan Generation}
In the final stage, the generated Problem PDDL is paired with a predefined Domain PDDL and passed to an off-the-shelf symbolic planner (e.g., Fast Downward). Because the solver operates on rigorous symbolic logic, any successfully generated plan is guaranteed to be feasible within the defined domain. 

Powered by a lightweight LLM, the Fast System resolves the majority of routine tasks in a single pass with minimal latency. However, if the LLM commits a semantic error during the extraction phase—such as omitting a critical implicit state—the symbolic planner will fail to find a solution and flag the problem as unsolvable. This failure state serves as the trigger to activate our \textbf{Slow System} for deeper reflection and targeted repair.

\section{DUPLEX: Agentic Dual-System Architecture With Reflection}
\label{sec:duplex}

\subsection{Motivation}
While the Fast System excels at rapidly processing nominal scenarios with minimal computational overhead, its feed-forward nature leaves it vulnerable to the inherent complexities of unstructured real-world environments. Even when confined to Information Extraction (IE) guided by the given domain PDDL file, lightweight LLMs may still  hallucinate non-existing predicates or objects or omit types for objects or omit critical implicit states (e.g., failing to notice that an
microwave oven is closed when it is running). Thus planning fails.

To maximize success rates, we draw inspiration from dual-process theory in cognitive science to propose the the \textbf{DU}al-System \textbf{P}lanning via \textbf{L}LM-Driven Information \textbf{EX}traction (\textbf{DUPLEX}) architecture. DUPLEX integrates the Fast System with an analytical Slow System. This formulation transforms the static pipeline into an agentic, closed-loop framework capable of iterative replanning and self-correction.

\subsection{The Slow System: Agentic Reflection}

The Slow System serves as the analytical engine. To preserve overall system efficiency, it remains dormant during successful Fast System operations and is activated exclusively upon planning failures. 

When triggered, the Slow System engages a high-capacity reasoning LLM (e.g., GPT-4o) to act as a diagnostic agent. By parsing the failure diagnostics from the symbolic planner, the Slow System performs root-cause analysis and executes targeted repairs by modifying the Problem PDDL to reflect corrected states.

\subsection{Hierarchical Validation and Self-Correction}
To systematically mitigate LLM hallucinations, DUPLEX employs a defense-in-depth self-correction pipeline. We categorize potential errors into three hierarchical levels, each handled by a dedicated validation and correction mechanism:

\subsubsection{Level 1: Rule-based Validation and Correction (Fast System)}
The initial layer enforces structural integrity right after the Fast System's IE phase. This module verifies the extracted information, checking for missing mandatory keys or malformed syntax (for instance, missing object types recoverable via the domain PDDL file). By employing a rule-based validation script, this module automatically detects and rectifies these elementary faults, thereby preempting trivial compilation failures.

\subsubsection{Level 2: "Semantic" Validation and Correction (Slow System)}
This advanced validation layer mitigates lightweight LLM-induced semantic anomalies, specifically out-of-vocabulary hallucinations (e.g., generating fictitious objects, types, or predicates) and omission of critical entities specified in the task description. The Slow System acts as an error-recovery mechanism, it intercepts compilation faults, drives larger LLM to cross-references the problematic output against both the domain PDDL and the original task description, and systematically resolves this type of errors.

\subsubsection{Level 3: Planner Diagnostics and Logical Reflection (Slow System)}
The most challenging errors occur when the generated PDDL is syntactically and semantically flawless, yet logically incomplete. In these cases, the planner executes successfully but fails to find a feasible solution, returning a \textsc{SEARCH\_FAIL} diagnostic. This indicates that the goal is unreachable under current initial state—most commonly due to omitted implicit states (e.g., failing to notice that a microwave oven is closed when it is working). 

Rather than failing blindly, the Slow System leverages these algorithmic diagnostics as an explicit reflection signal. It analyzes causal dependencies, infers implicit states, and iteratively fixes the Problem PDDL. This solver-driven reflection enables the agent to discover and correct its own cognitive blind spots, ensuring robust long-horizon planning.
\section{Experiments}\label{experiment}

\subsection{Evaluation Metrics}
We use \textbf{Success Rate (SR)} as the primary evaluation metric:
\begin{equation}
    SR = \frac{N_{solved}}{N} \times 100\%
\end{equation}
A task is counted as successful only if the generated plan is valid: the planner finds a solution within the time limit, and the plan is verified as executable and goal-reaching. All other cases are counted as failures.

\subsection{Benchmark Problems}
We evaluate our method across two distinct benchmark categories: classical International Planning Competition (IPC) domains \cite{vallati2015ipc} and embodied household planning domains \cite{liu2024delta}. Collectively, these benchmarks span a broad spectrum of planning challenges, ranging from combinatorial tasks governed by rigid logical rules to embodied scenarios requiring commonsense semantic reasoning.

\subsubsection{Classical IPC Domains}
The IPC domains evaluate an agent's ability to handle long-horizon dependencies, strict causal constraints, and complex symbolic reasoning. We select eight representative domains, evaluating 20 test tasks for each:
\begin{description}
    \item[Barman:] Multi-step liquid mixing and container manipulation, requiring intricate action sequencing.
    \item[Blocksworld:] A fundamental stacking domain governed by spatial relational constraints (e.g., \textit{On}, \textit{Clear}).
    \item[Floortile:] A grid-routing and painting task constrained by geometric movement limitations.
    \item[Grippers:] A multi-room transportation task testing multi-object coordination and resource allocation.
    \item[Storage:] Crate manipulation subject to spatial connectivity and hoist constraints.
    \item[Termes:] A construction domain blending spatial navigation with hierarchical block building.
    \item[Tyreworld:] A mechanical repair task with strict causal action ordering.
    \item[Visitall:] A grid-coverage problem requiring the agent to navigate to all specified locations.
\end{description}

\subsubsection{Embodied Household Domains}
Following the DELTA long-horizon planning framework \cite{liu2024delta}, these domains capture complex task execution in indoor robotic environments. We evaluate four task domains using three representative indoor instances each (12 instances total). Every task is executed 50 times to report the average success rate (SR):
\begin{description}
    \item[PC (PC Assembly):] A hardware assembly task constrained by strong multi-object dependencies.
    \item[Dining (Table Setting):] A relational arrangement task requiring multi-object transport and spatial reasoning.
    \item[Cleaning (Household Cleaning):] A long-horizon maintenance task involving disposal, tool utilization, and resource replenishment.
    \item[Office (Home Office Setup):] An organization task governed by container constraints and logical object placement.
\end{description}

\subsection{Baselines}
\begin{enumerate}
    \item \textbf{LLM-as-Planner:} The LLM generates the action sequence end-to-end.
    
    \item \textbf{LLM+P:} The \emph{translate-then-solve} approach from \cite{liu2023llm+}. The LLM first translates the problem into a PDDL file, which is then solved by a classical planner.
\end{enumerate}

\subsection{Experimental Setup}
To ensure deterministic generation, we apply greedy decoding (Top-p = 1) across all language models. GPT-4o serves as the foundational model for all baselines and domains. In contrast, our proposed DUPLEX framework employs a dual-system architecture: Qwen3-8B acts in the Fast System, while GPT-4o acts in the Slow System.

For methods integrating an external classical planner (LLM+P and DUPLEX), we utilize Fast Downward. The planner configurations are domain-specific: classical IPC domains use the LAMA heuristic (\texttt{lama-first}), whereas embodied household domains employ \texttt{seq-opt-lmcut}. As structured in the household benchmark, evaluating three scene instances across four domains for 50 trials each yields a comprehensive suite of 600 execution trials.

We enforce strict time limits of 60 seconds for LLM generation and 500 seconds for classical planning. A task is recorded as a failure if it exceeds either time budget, yields an unparsable or unsolvable PDDL formulation, or fails the final plan validation. To maintain a fair comparison, all evaluated methods share identical few-shot example counts and selection strategies. All local LLM inference is executed on a Linux server equipped with a single NVIDIA RTX 4090 GPU and 32 GB of RAM.

\section{Results And Discussion}
\subsection{Main Results: Success Rate Analysis}

Table \ref{tab:ipc_results} summarizes the IPC benchmark results. LLM-as-Planner struggles in domains with strict precondition--effect dependencies and long-horizon structures, frequently generating outputs that appear plausible but fail logically. LLM+P significantly improves upon this by integrating symbolic search. Our method, DUPLEX, drives performance even further by restricting the LLM to schema-based extraction and employing an agentic, dual-system reflexion architecture. Overall, DUPLEX achieves an average success rate of 97.5\% across IPC domains, outperforming LLM+P (71.9\%) by 25.6\% and vastly exceeding LLM-as-Planner (11.9\%).

The benefits of solver-driven reflexion are most pronounced in domains where failures stem from missing facts or brittle grounding rather than search complexity. For instance, DUPLEX increases the success rate in \textsc{Floortile} from 30.0\% (Fast System only) to 85.0\% (Full System), and in \textsc{Visitall} from 40.0\% to 95.0\%. These gains demonstrate the Slow System's ability to reliably diagnose and repair flawed state descriptions when the planner encounters infeasibility. Conversely, in domains with stable semantic mappings like \textsc{Barman} and \textsc{Grippers}, the Fast System alone suffices.

Table \ref{tab:household_results} details the results for long-horizon household tasks. Compared to the IPC benchmark, these tasks feature richer semantics and heavily underspecified initial conditions, making them highly sensitive to missing objects, incorrect relations, and implicit constraints. DUPLEX overcomes these challenges using schema constraints paired with planner-driven repair. Ultimately, it achieves an 83.5\% average success rate, substantially outperforming both LLM-as-Planner (27.2\%) and LLM+P (20.0\%).

\begin{table}[h]
\centering
\caption{Success Rates (\%) on IPC Domains}
\label{tab:ipc_results}
\small
\renewcommand{\arraystretch}{1.2}
\begin{tabularx}{\columnwidth}{@{} l C C C C @{} }
\toprule
\multirow{2}{*}{\textbf{Domain}} & \textbf{LLM-as-Planner} & \textbf{LLM+P} & \textbf{DUPLEX Fast System} & \textbf{DUPLEX Full System} \\
 & (GPT-4o) & (GPT-4o) & \textbf{(Qwen3-8B)} & \textbf{(Qwen3-8B + GPT-4o)} \\
\midrule
Barman      & 0.0  & 100.0 & 100.0 & \textbf{100.0} \\
Blocksworld & 30.0 & 90.0  & 100.0 & \textbf{100.0} \\
Floortile   & 0.0  & 15.0  & 30.0  & \textbf{85.0} \\
Grippers    & 50.0 & 100.0 & 100.0 & \textbf{100.0} \\
Storage     & 0.0  & 85.0  & 100.0 & \textbf{100.0} \\
Termes      & 0.0  & 90.0  & 100.0 & \textbf{100.0} \\
Tyreworld   & 15.0 & 90.0  & 100.0 & \textbf{100.0} \\
Visitall    & 0.0  & 5.0   & 40.0  & \textbf{95.0} \\
\midrule
\textit{Average} & \textit{11.9} & \textit{71.9} & \textit{\textbf{83.8}} & \textit{\textbf{97.5}} \\
\bottomrule
\end{tabularx}
\end{table}
\begin{table}[h]
\centering
\caption{Success Rates (\%) on Household Domains}
\label{tab:household_results}
\small
\renewcommand{\arraystretch}{1.2}
\begin{tabularx}{\columnwidth}{@{} l C C C C @{} }
\toprule
\multirow{2}{*}{\textbf{Domain}} & \textbf{LLM-as-Planner} & \textbf{LLM+P} & \textbf{DUPLEX Fast System} & \textbf{DUPLEX Full System} \\
& (GPT-4o) & (GPT-4o) & \textbf{(Qwen3-8B)} & \textbf{(Qwen3-8B + GPT-4o}\textbf{)} \\
\midrule
PC Assembly  & 70.0  & 76.0  & \textbf{76.7} & \textbf{96.7} \\
Dining Setup & 38.7  & 4.0   & \textbf{70.0} & \textbf{98.0} \\
Cleaning     & 0.0   & 0.0   & \textbf{30.7} & \textbf{70.7} \\
Office       & 0.0   & 0.0   & \textbf{26.0} & \textbf{68.7} \\
\midrule
\textit{Average} & \textit{27.2} & \textit{20.0} & \textit{\textbf{50.9}} & \textit{\textbf{83.5}} \\
\bottomrule
\end{tabularx}
\end{table}

\subsection{Ablation Study: The Impact of Reflexion}

To isolate the contribution of our dual-system architecture, we ablate the Slow System (the reflexion module). As shown in Table \ref{tab:ablation}, relying solely on the feed-forward Fast System yields success rates of 83.8\% on the IPC benchmark and 50.9\% on household tasks. 
\begin{table}[h]
\centering
\caption{Ablation Study: Effectiveness of the Slow System}
\label{tab:ablation}
\small
\renewcommand{\arraystretch}{1.2}
\begin{tabularx}{\columnwidth}{@{} X C C @{}}
\toprule
\multirow{2}{*}{\textbf{Methods}} & \multicolumn{2}{c}{\textbf{Success Rate (\%)}} \\
\cmidrule(lr){2-3}
 & IPC & Household \\
\midrule
\textbf{Full System} & \textbf{97.5} & \textbf{83.5} \\
Fast System Only & 83.8 & 50.9 \\
\bottomrule
\end{tabularx}
\end{table}
These results yield two key insights. First, the Fast System alone establishes a strong baseline, confirming that schema-guided information extraction coupled with deterministic PDDL mapping is inherently more robust than LLM-as-Planner and LLM+P. Second, single-pass extraction remains insufficient for underspecified, complex environments. Activating the Slow System provides absolute improvements of 13.7\% and 32.6\% on the IPC and household benchmarks, respectively. The substantial gain in the household domains underscores the necessity of the planner-validator closed loop, which effectively diagnoses and repairs the grounding errors, missing facts, and semantic inconsistencies that a single-pass extractor inevitably overlooks.

\section{Conclusion}\label{conclusion}

This paper introduces DUPLEX, an agentic dual-system neuro-symbolic architecture that redefines the role of LLMs in long-horizon task planning. Rather than functioning as end-to-end planners prone to logical hallucinations, LLMs in our framework serve as schema-guided information extractors coupled with symbolic solvers. Our evaluations demonstrate that combining structured extraction with failure-triggered agentic reflexion significantly improves planning reliability and successfully resolves implicit state violations in complex embodied domains.

Despite these advancements, DUPLEX currently assumes access to manually authored domain PDDL files. Furthermore, the iterative reflexion loop introduces latency that may bottleneck high-frequency real-time control, and relying exclusively on textual extraction leaves the system vulnerable to referential ambiguity in visually rich environments. Future work will focus on automatically learning action schemas from multimodal execution traces and optimizing the reflection cycle to enable more responsive, closed-loop robotic deployment in open-world settings.

\footnotesize
\bibliographystyle{IEEEtran}
\bibliography{refs}

@article{mcdermott1998pddl,
  title     = {PDDL---The Planning Domain Definition Language},
  author    = {McDermott, Drew and Ghallab, Malik and Howe, Adele and Knoblock, Craig and Ram, Ashwin and Veloso, Manuela and Weld, Daniel and Wilkins, David},
  year      = {1998},
  url       = {https://homepages.inf.ed.ac.uk/mfourman/tools/propplan/pddl.pdf}
}

@article{fox2003pddl21,
  title   = {{PDDL}2.1: An Extension to {PDDL} for Expressing Temporal Planning Domains},
  author  = {Fox, Maria and Long, Derek},
  journal = {Journal of Artificial Intelligence Research},
  volume  = {20},
  pages   = {61--124},
  year    = {2003},
  doi     = {10.1613/jair.1129},
  url     = {https://jair.org/index.php/jair/article/view/10357}
}

@article{helmert2006fast,
  title   = {The Fast Downward Planning System},
  author  = {Helmert, Malte},
  journal = {Journal of Artificial Intelligence Research},
  volume  = {26},
  pages   = {191--246},
  year    = {2006},
  doi     = {10.1613/jair.1705},
  url     = {https://jair.org/index.php/jair/article/view/10390}
}

@inproceedings{howey2004val,
  title        = {{VAL}: Automatic Plan Validation, Continuous Effects and Mixed Initiative Planning Using {PDDL}},
  author       = {Howey, Richard and Long, Derek and Fox, Maria},
  booktitle    = {16th IEEE International Conference on Tools with Artificial Intelligence},
  pages        = {294--301},
  year         = {2004},
  organization = {IEEE},
  doi          = {10.1109/ICTAI.2004.50}
}

@article{hoffmann2001ff,
  title   = {The {FF} Planning System: Fast Plan Generation Through Heuristic Search},
  author  = {Hoffmann, J{"o}rg and Nebel, Bernhard},
  journal = {Journal of Artificial Intelligence Research},
  volume  = {14},
  pages   = {253--302},
  year    = {2001},
  doi     = {10.1613/jair.855},
  url     = {https://jair.org/index.php/jair/article/view/10374}
}

@book{ghallab2004automatedplanning,
  title     = {Automated Planning: Theory and Practice},
  author    = {Ghallab, Malik and Nau, Dana and Traverso, Paolo},
  publisher = {Morgan Kaufmann},
  year      = {2004},
  isbn      = {9781558608566}
}

@article{brown2020language,
  title   = {Language Models Are Few-Shot Learners},
  author  = {Brown, Tom B. and Mann, Benjamin and Ryder, Nick and Subbiah, Melanie and Kaplan, Jared and Dhariwal, Prafulla and Neelakantan, Arvind and Shyam, Pranav and Sastry, Girish and Askell, Amanda and others},
  journal = {Advances in Neural Information Processing Systems},
  volume  = {33},
  pages   = {1877--1901},
  year    = {2020},
  url     = {https://arxiv.org/abs/2005.14165}
}

@article{ouyang2022instructgpt,
  title   = {Training Language Models to Follow Instructions with Human Feedback},
  author  = {Ouyang, Long and Wu, Jeffrey and Jiang, Xu and Almeida, Diogo and Wainwright, Carroll and Mishkin, Pamela and Zhang, Chong and Agarwal, Sandhini and Slama, Katarina and Ray, Alex and others},
  journal = {arXiv preprint arXiv:2203.02155},
  year    = {2022},
  doi     = {10.48550/arXiv.2203.02155},
  url     = {https://arxiv.org/abs/2203.02155}
}

@article{shinn2023reflexion,
  title   = {Reflexion: Language Agents with Verbal Reinforcement Learning},
  author  = {Shinn, Noah and Cassano, Federico and Berman, Edward and Gopinath, Ashwin and Narasimhan, Karthik and Yao, Shunyu},
  journal = {arXiv preprint arXiv:2303.11366},
  year    = {2023},
  doi     = {10.48550/arXiv.2303.11366},
  url     = {https://arxiv.org/abs/2303.11366}
}

@article{yao2023tot,
  title   = {Tree of Thoughts: Deliberate Problem Solving with Large Language Models},
  author  = {Yao, Shunyu and Yu, Dian and Zhao, Jeffrey and Shafran, Izhak and Griffiths, Thomas L. and Cao, Yuan and Narasimhan, Karthik},
  journal = {arXiv preprint arXiv:2305.10601},
  year    = {2023},
  doi     = {10.48550/arXiv.2305.10601},
  url     = {https://arxiv.org/abs/2305.10601}
}

@article{madaan2023selfrefine,
  title   = {Self-Refine: Iterative Refinement with Self-Feedback},
  author  = {Madaan, Aman and Tandon, Niket and Gupta, Prakhar and Hallinan, Skyler and Gao, Luyu and Wiegreffe, Sarah and Alon, Uri and Dziri, Nouha and Prabhumoye, Shrimai and others},
  journal = {arXiv preprint arXiv:2303.17651},
  year    = {2023},
  doi     = {10.48550/arXiv.2303.17651},
  url     = {https://arxiv.org/abs/2303.17651}
}

@article{dhuliawala2023cove,
  title   = {Chain-of-Verification Reduces Hallucination in Large Language Models},
  author  = {Dhuliawala, Shehzaad and Komeili, Mojtaba and Xu, Jing and Raileanu, Roberta and Li, Xian and Celikyilmaz, Asli and Weston, Jason},
  journal = {arXiv preprint arXiv:2309.11495},
  year    = {2023},
  doi     = {10.48550/arXiv.2309.11495},
  url     = {https://arxiv.org/abs/2309.11495}
}

@article{openai2023gpt4,
  title   = {GPT-4 Technical Report},
  author  = {{OpenAI}},
  journal = {arXiv preprint arXiv:2303.08774},
  year    = {2023},
  doi     = {10.48550/arXiv.2303.08774},
  url     = {https://arxiv.org/abs/2303.08774}
}

@article{valmeekam2022large,
  title   = {Large Language Models Still Can't Plan (A Benchmark for LLMs on Planning and Reasoning about Change)},
  author  = {Valmeekam, Karthik and Olmo, Alberto and Sreedharan, Sarath and Kambhampati, Subbarao},
  journal = {arXiv preprint arXiv:2206.10498},
  year    = {2022},
  doi     = {10.48550/arXiv.2206.10498},
  url     = {https://arxiv.org/abs/2206.10498}
}

@inproceedings{huang2022language,
  title     = {Language Models as Zero-Shot Planners: Extracting Actionable Knowledge for Embodied Agents},
  author    = {Huang, Wenlong and Abbeel, Pieter and Pathak, Deepak and Mordatch, Igor},
  booktitle = {International Conference on Machine Learning (ICML)},
  pages     = {9118--9147},
  year      = {2022},
  url       = {https://proceedings.mlr.press/v162/huang22a.html}
}

@article{ahn2022i,
  title   = {Do as {I} Can, Not as {I} Say: Grounding Language in Robotic Affordances},
  author  = {Ahn, Michael and Brohan, Anthony and Brown, Noah and Chebotar, Yevgen and Cortes, Omar and David, Byron and Finn, Chelsea and Fu, Chuyuan and Gopalakrishnan, Keerthana and Hausman, Karol and others},
  journal = {arXiv preprint arXiv:2204.01691},
  year    = {2022},
  doi     = {10.48550/arXiv.2204.01691},
  url     = {https://arxiv.org/abs/2204.01691}
}

@article{liu2023llm+,
  title   = {{LLM}+{P}: Empowering Large Language Models with Optimal Planning Proficiency},
  author  = {Liu, Bo and Jiang, Yuqian and Zhang, Xiaohan and Liu, Qiang and Zhang, Shiqi and Biswas, Joydeep and Stone, Peter},
  journal = {arXiv preprint arXiv:2304.11477},
  year    = {2023},
  doi     = {10.48550/arXiv.2304.11477},
  url     = {https://arxiv.org/abs/2304.11477}
}

@inproceedings{song2023llm,
  title     = {{LLM}-Planner: Few-Shot Grounded Planning for Embodied Agents with Large Language Models},
  author    = {Song, Chan Hee and Wu, Jiaman and Washington, Clayton and Sadler, Brian M. and Chao, Wei-Lun and Su, Yu},
  booktitle = {Proceedings of the IEEE/CVF International Conference on Computer Vision (ICCV)},
  pages     = {2998--3009},
  year      = {2023},
  doi       = {10.1109/ICCV51070.2023.00279}
}

@article{pallagani2023understanding,
  title   = {Understanding the Capabilities of Large Language Models for Automated Planning},
  author  = {Pallagani, Vishal and Muppasani, Bharath and Murugesan, Keerthiram and Rossi, Francesca and Srivastava, Biplav and Horesh, Lior and Fabiano, Francesco and Loreggia, Andrea},
  journal = {arXiv preprint arXiv:2305.16151},
  year    = {2023},
  doi     = {10.48550/arXiv.2305.16151},
  url     = {https://arxiv.org/abs/2305.16151}
}

@inproceedings{liang2023code,
  title     = {Code as Policies: Language Model Programs for Embodied Control},
  author    = {Liang, Jacky and Huang, Wenlong and Xia, Fei and Xu, Peng and Hausman, Karol and Ichter, Brian and Florence, Pete and Zeng, Andy},
  booktitle = {2023 IEEE International Conference on Robotics and Automation (ICRA)},
  pages     = {9493--9500},
  year      = {2023},
  doi       = {10.1109/ICRA48891.2023.10160943},
  url       = {https://arxiv.org/abs/2209.07753}
}

@inproceedings{shah2023lm,
  title     = {{LM}-Nav: Robotic Navigation with Large Pre-Trained Models of Language, Vision, and Action},
  author    = {Shah, Dhruv and Osi{\'n}ski, B{\l}a{\.{z}}ej and Levine, Sergey and others},
  booktitle = {Conference on Robot Learning (CoRL)},
  pages     = {492--504},
  year      = {2023},
  url       = {https://proceedings.mlr.press/v205/shah23a.html}
}

@article{wang2023voyager,
  title   = {Voyager: An Open-Ended Embodied Agent with Large Language Models},
  author  = {Wang, Guanzhi and Xie, Yuqi and Jiang, Yunfan and Mandlekar, Ajay and Xiao, Chaowei and Zhu, Yuke and Fan, Linxi and Anandkumar, Anima},
  journal = {arXiv preprint arXiv:2305.16291},
  year    = {2023},
  doi     = {10.48550/arXiv.2305.16291},
  url     = {https://arxiv.org/abs/2305.16291}
}

@article{driess2023palme,
  title   = {{PaLM}-{E}: An Embodied Multimodal Language Model},
  author  = {Driess, Danny and Xia, Fei and Sajjadi, Mehdi S. M. and Lynch, Corey and Chowdhery, Aakanksha and Ichter, Brian and Wahid, Ayzaan and Tompson, Jonathan and Vuong, Quan and others},
  journal = {arXiv preprint arXiv:2303.03378},
  year    = {2023},
  doi     = {10.48550/arXiv.2303.03378},
  url     = {https://arxiv.org/abs/2303.03378}
}

@article{xu2024largelanguagemodelsgenerative,
      title={Large Language Models for Generative Information Extraction: A Survey}, 
      author={Derong Xu and Wei Chen and Wenjun Peng and Chao Zhang and Tong Xu and Xiangyu Zhao and Xian Wu and Yefeng Zheng and Yang Wang and Enhong Chen},
      year={2024},
      eprint={2312.17617},
      archivePrefix={arXiv},
      primaryClass={cs.CL},
      url={https://arxiv.org/abs/2312.17617}, 
}

@article{wei2024chatiezeroshotinformationextraction,
      title={ChatIE: Zero-Shot Information Extraction via Chatting with ChatGPT}, 
      author={Xiang Wei and Xingyu Cui and Ning Cheng and Xiaobin Wang and Xin Zhang and Shen Huang and Pengjun Xie and Jinan Xu and Yufeng Chen and Meishan Zhang and Yong Jiang and Wenjuan Han},
      year={2024},
      eprint={2302.10205},
      archivePrefix={arXiv},
      primaryClass={cs.CL},
      url={https://arxiv.org/abs/2302.10205}, 
}

@article{shinn2023reflexionlanguageagentsverbal,
      title={Reflexion: Language Agents with Verbal Reinforcement Learning}, 
      author={Noah Shinn and Federico Cassano and Edward Berman and Ashwin Gopinath and Karthik Narasimhan and Shunyu Yao},
      year={2023},
      eprint={2303.11366},
      archivePrefix={arXiv},
      primaryClass={cs.AI},
      url={https://arxiv.org/abs/2303.11366}, 
}

@article{madaan2023selfrefineiterativerefinementselffeedback,
      title={Self-Refine: Iterative Refinement with Self-Feedback}, 
      author={Aman Madaan and Niket Tandon and Prakhar Gupta and Skyler Hallinan and Luyu Gao and Sarah Wiegreffe and Uri Alon and Nouha Dziri and Shrimai Prabhumoye and Yiming Yang and Shashank Gupta and Bodhisattwa Prasad Majumder and Katherine Hermann and Sean Welleck and Amir Yazdanbakhsh and Peter Clark},
      year={2023},
      eprint={2303.17651},
      archivePrefix={arXiv},
      primaryClass={cs.CL},
      url={https://arxiv.org/abs/2303.17651}, 
}

@article{vallati2015ipc,
  title={The 2014 International Planning Competition: Progress and Trends},
  author={Vallati, Mauro and Chrpa, Luk{\'a}{\v{s}} and Grze{\'s}, Marek and McCluskey, Thomas Leo and Roberts, Mark and Sanner, Scott},
  journal={AI Magazine},
  volume={36},
  number={3},
  pages={90--98},
  year={2015},
  publisher={Association for the Advancement of Artificial Intelligence (AAAI)}
}

@inproceedings{liu2024delta,
  title={DELTA: Decomposed Efficient Long-Term Robot Task Planning using Large Language Models},
  author={Liu, Yuchen and Palmieri, Luigi and Koch, Sebastian and Georgievski, Ilche and Aiello, Marco},
  booktitle={IEEE International Conference on Robotics and Automation (ICRA)},
  year={2025},
  publisher={IEEE}
}

\end{document}